\documentclass[conference]{IEEEtran}
\IEEEoverridecommandlockouts
\usepackage{float}
\usepackage{cite}
\usepackage{amsmath,amssymb,amsfonts}
\usepackage{graphicx}
\usepackage{textcomp}
\usepackage{amsmath}
\usepackage{algorithm}
\usepackage[noend]{algpseudocode}
\usepackage{caption}
\usepackage{listings}
\usepackage{blindtext}
\usepackage{array}
\usepackage{booktabs}
\usepackage{makecell}

\newcommand{\lane}{$lane$}
\newcommand{\lanes}{$lanes$}

\makeatletter
\def\BState{\State\hskip-\ALG@thistlm}
\makeatother

\usepackage[caption = false]{subfig}

\usepackage{xcolor}
\def\BibTeX{{\rm B\kern-.05em{\sc i\kern-.025em b}\kern-.08em
    T\kern-.1667em\lower.7ex\hbox{E}\kern-.125emX}}
\begin{document}

\title{Efficiency and Scalability of\\ Multi-Lane Capsule Networks (MLCN)\\
}

\author{\IEEEauthorblockN{Vanderson M. do Rosario}
\IEEEauthorblockA{\textit{Institute of Computing} \\
\textit{Unicamp}\\
Campinas, Brasil \\
vanderson.rosario@ic.unicamp.br}
\and
\IEEEauthorblockN{Mauricio Breternitz Jr.}
\IEEEauthorblockA{\textit{ISTAR-IUL} \\
\textit{Lisbon University Institute ISCTE-IUL}\\
Lisbon, Portugal \\
mbjrz@iscte-iul.pt}
\and
\IEEEauthorblockN{Edson Borin}
\IEEEauthorblockA{\textit{Institute of Computing} \\
\textit{Unicamp}\\
Campinas, Brasil \\
edson@ic.unicamp.br}

\thanks{This work was supported in part by  CAPES/Brasil (Finance  Code  001), by CNPq (313012/2017-2), and by Fapesp (CCES 2013/08293-7).  We would like to thank Google Cloud Platform for a grant to run our experiments.}

\thanks{V. M. do Rosario, is a Ph.D. Candidate at Institute of Computing, Unicamp, Brazil. (e-mail: vanderson.rosario@ic.unicamp.br).}

\thanks{E. Borin, is an Associate Professor at Institute of Computing, Unicamp, Brazil. (e-mail: edson@ic.unicamp.br).}

\thanks{M. Breternitz, Jr., is an Invited Associate Professor and Principal Investigator at Lisbon University Institute ISCTE-IUL and ISTAR-IUL, Portugal. (e-mail: mbjrz@iscte-iul.pt).}
}

\maketitle

\begin{abstract}
Some Deep Neural Networks (DNN) have what we call lanes, or they can be reorganized as such. Lanes are paths in the network which are data-independent and typically learn different features or add resilience to the network. Given their data-independence, lanes are amenable for parallel processing. The Multi-lane CapsNet (MLCN) is a proposed reorganization  of the Capsule Network which is shown to achieve better accuracy while bringing  highly-parallel lanes. However, the efficiency and scalability of  MLCN had not been systematically examined. In this work, we study the MLCN network with multiple GPUs finding that it is 2x more efficient than the original CapsNet when using model-parallelism. Further, we present the load balancing problem of distributing heterogeneous lanes in homogeneous or heterogeneous accelerators and show that a simple greedy heuristic can be almost 50\% faster than a na\"ive random approach.
\end{abstract}


\section{Introduction}

Several approaches to the distributed model parallelization of Deep Neural Networks (DNN) have concentrated in their depth \cite{huang2018gpipe,mehta2018high,ben2018demystifying}, but DNNs can also be organized in a way to be parallelized in their width \cite{jia2018beyond}. The DNN architecture may be organized into distinct neural network \lanes\cite{MLCN-SPL}. This creates separable and resource efficient data-independent paths in the network that can be used to learn different features or add resilience to the network. Examples of neural networks with \lanes\ are the Google Inception \cite{chollet2017xception,szegedy2017inception} and the Multi-lane Capsule Network (MLCN) \cite{MLCN-SPL}. As these \lanes\ are data-independent they can be  (1) processed in parallel and (2) specialized for distinct computational targets (CPUs, GPU, FPGAs, and cloud), as well as resource-constrained mobile and IoT targets, leading to opportunities and challenges. Recently, our research focus was on Multi-Lane Capsule Networks (MLCN), which are a separable and resource efficient organization of Capsule Networks (CapsNet) that allows parallel processing while achieving high accuracy at a reduced cost. Table \ref{tab:comp} shows  results from MLCN in comparison with the baseline CapsNet. With a similar number of parameters, MLCN achieves similar accuracy but with a significant speedup stemming from the \lane\ organization. Initial experiments were performed in single GPU environments but, with  highly-parallel \lanes\, it is interesting to explore how MLCN scales with more GPUs. Here we present a first comprehensive study of the scalability and efficiency of MLCN for multi-GPU systems.

\begin{table}[!h]
	\centering
	\small
	\caption{Comparison between Baseline CapsNet and MLCN.}
	{\setlength{\tabcolsep}{4pt}
	\begin{tabular}[t]{lccccc}
	%	\begin{tabular}[t]{p{0.16\linewidth}p{0.06\linewidth}p{0.065\linewidth}p{0.09\linewidth}p{0.22\linewidth}p{0.11\linewidth}}
		\toprule
		\makecell{Network/set} & \makecell{\# of\\\lanes} & \makecell{\lane's\\Width} & \makecell{Params.} & \makecell{Train Time\\(sec./epoch)} & \makecell{Accuracy}\\
		\midrule
		\textbf{Cifar10:} &&&&&\\
		Baseline &  - & - & 11k & 240 & 66.36\%\\
		Mlcn2 &  4 & 4 & 5k & 53 & 69.05\%\\
		Mlcn2 &   32 & 2 & 14k & 204 & 75.18\%\\
		\\
		\multicolumn{2}{l}{\textbf{Fashion-MNIST:}} &&&&\\
		Baseline &  - & - & 8k & 220 & 91.30\%\\
		Mlcn2 & 2 & 4 & 3.6k & 20 & 91.01\%\\
		Mlcn2 &  8 & 4 & 10.6k & 92 & 92.63\%\\
		\bottomrule
		\label{tab:comp}
	\end{tabular}}
\end{table}

Moreover, the \lanes\ do not necessarily need to have the same sizes or shapes and may perhaps even learn different features of the given task. This implies that each distinct \lane\ may be better suitable for a distinct HW substrate. Further, each \lane\ may tolerate different impacts from various optimizations (such as quantization). Thus, given a set of \lanes, $L$, and a set of hardware (HW), $H$, there is an optimal pair $(l, h)$ for $l \in L$ and $h \in H$ and an optimal sequence of \lane\ optimizations for each pair $(l, d)$ of \lane\ and HW.
        
In this work, we describe and present this lane-hardware matching problem for homogeneous or heterogeneous accelerator scenarios. We  also show that a simple greedy heuristic can be almost 50\% faster than a random na\"ive approach. 

The main contributions of this work are:

\begin{itemize}
    \item We present a first comprehensive analysis of the efficiency and scalability of MLCN showing its advantages over the data-parallelism-limited approach of the original CapsNet.
    \item We define the load balancing problem of distributing heterogeneous \lanes\ in heterogeneous hardware.
    \item We present a greedy heuristic to solve the lane-hardware match problem showing that it is superior to a na\"ive approach.
\end{itemize}

This paper is organized as follows: 
Section~\ref{sec:relatedwork} presents the state-of-art in Capsule Networks and DNN parallelization; 
Section~\ref{sec:mlcn} describes the Multi-Lane Capsule Network (MLCN) and discusses how it can be parallelized; 
Section~\ref{sec:heteproblem} further discusses the heterogeneous distribution problem and presents a heuristic approach to it; 
finally, Section~\ref{sec:expsetup} and \ref{sec:expresult} shows the experimental setup and the experimental results, and Section~\ref{sec:conclusion} presents our conclusions.

\section{Related Work}\label{sec:relatedwork}

\subsection{Capsule Network}

The Convolutional Neural Network (CNN) is a class of DNN which is commonly used when working with images. CNNs have already achieved state of art results in tasks such as image and video recognition, image classification and medical image analysis. However, these networks have difficulties with location invariance and loss of location information, e.g., one CNN which is able to recognize faces could also mistakenly recognize an image with eyes, mouth, and nose at random positions as a face, not understanding that there is an important spatial relationship between the composing elements. To address this problem, many different new DNN approaches were proposed, including the notion of capsules proposed by Hiton, Krizhevsky, and Wang in 2011 \cite{hinton2011transforming}.

To encode spatial relationship, Capsule Networks also known as CapsNets, do not work/represent neurons as simple scalars (as in regular CNNs), but as vectors. Later in 2017 an efficient and realistic training algorithm for such networks was proposed \cite{sabour2017dynamic}. The algorithm,  named Dynamic Routing, dynamically chooses activation paths between capsules from one layer to another, calculating the vectors from the next layer based on a mean from dynamically selected vectors from all previous layers. 

CapsNet \cite{sabour2017dynamic} produces a set of $N$ Primary Capsules (PCs) by applying two convolutional steps to the original image and splitting it in vectors. Each of these PCs (vectors), identified as $u_{i}$, is multiplied by a weight matrix $W_{i}$ and finally, a final set of capsules, the digit capsules, is created using the dynamic routing algorithm. Each of these digit capsule vectors represents one of the classes in the classification problem and the vector's length encodes the probability of the class. The digit capsule can also be used to reconstruct the image like an auto-encoder.

This network with the Dynamic Routing algorithm was shown to have some advantages such as a smaller necessary training set and location invariance. It also has some drawbacks such as slower execution and lower accuracy than CNNs. Since the initial publication, however, multiple improvements were proposed and the concept has been evolving. Shahroudnejad, Mohammadi, and Plataniotis \cite{shahroudnejad2018improved} presented an analysis of the explainability of CapsNet, showing that it has properties to help understand and explain its behavior. Jaiswal \textit{et al}. 
\cite{jaiswal2018capsulegan} used the CapsNet in a Generative Adversarial Network (GAN) and showed that it can achieve lower error rates than the simple CNN.
Ren and Lu  \cite{ren2018compositional} showed that CapsNet can be used for text classification and showed how to adapt the compositional coding mechanism to the CapsNet architecture. 
Jimenez-Sanchez, Albarqouni, and Mateus \cite{jimenez2018capsule} tested  CapsNet for Medical Imaging Data Challenges showing that it can achieve good performance even when having less trainable parameters than the tested counterpart CNNs.
Mobiny and Nguyen \cite{mobiny2018fast} tested the performance of CapsNet for lung cancer screening and showed that it could outperform CNNs mainly when the training set was small.
A similar result was achieved by Kim et al. in traffic speed prediction \cite{kim2018capsule} with CapsNet outperforming traditional CNNs approaches.
Mukhometzianov and Carrillo \cite{mukhometzianov2018capsnet} used  CapsNet with multiple image datasets and found that, although achieving good results, CapsNet still requires higher training times compared to other CNNs.
Canqun et al. \cite{xiang2018ms} proposed the Multi-Scale CapsNet (MS-CapsNet). They introduced a fixed division of the CapsNet network limited to three ``\lanes" (they did neither name or explore the division concept), each with a different number of convolutions. 
Also, recently developed,  the Path Capsule Networks by Amer and Maul \cite{path2019} (Path-Capsnet)  explore the parallelism of CapsNets by splitting the network such that each path or \lane\ is responsible for computing each digitcaps or a primary capsule entirely, unlike the computation of different dimensions/features in MLCN.

\section{Multi-lane CapsNets (MLCN)}\label{sec:mlcn}

\begin{figure}[h!]
    \centering
    \includegraphics[width=\columnwidth]{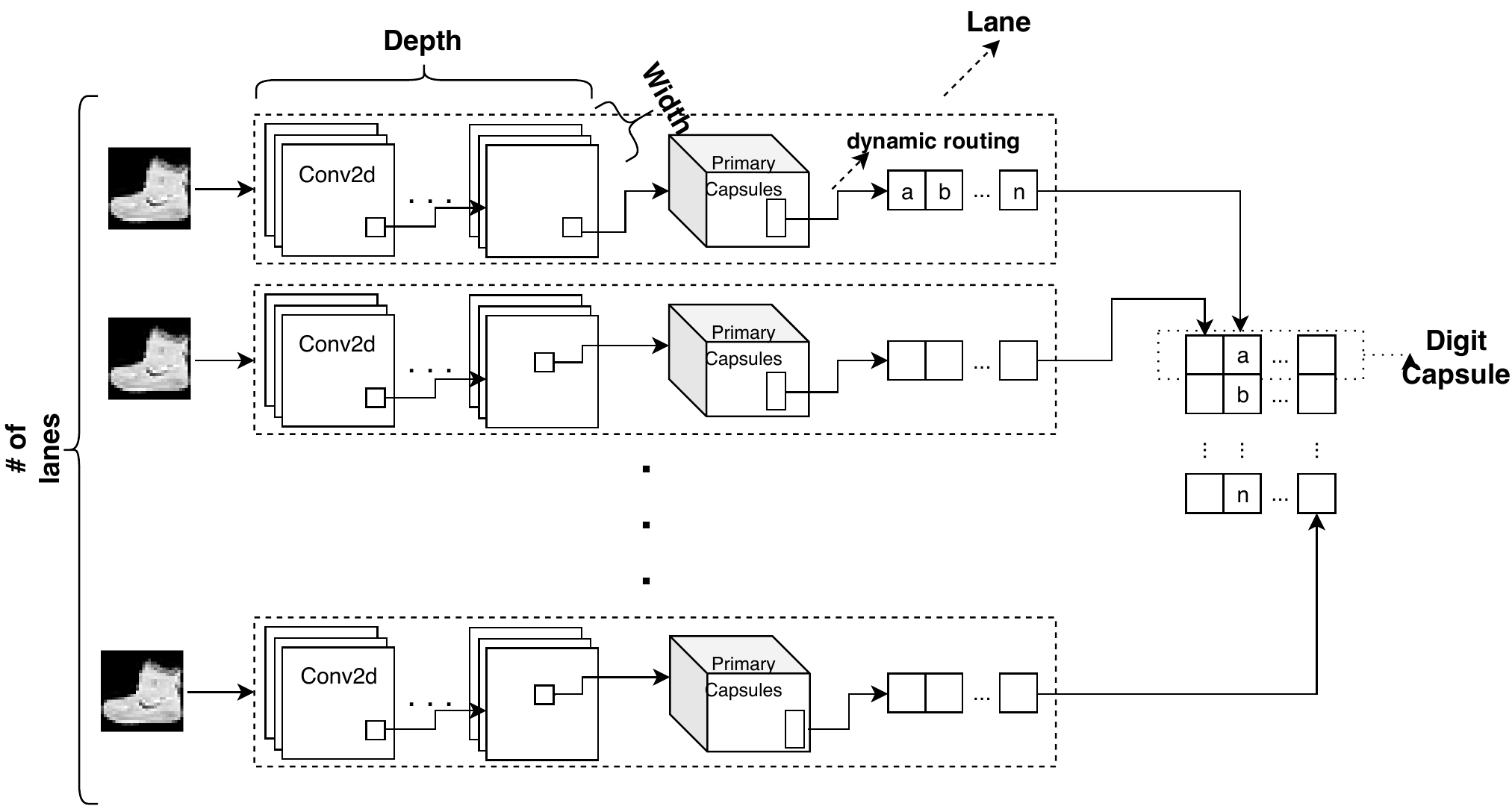}
    \caption{MLCN architecture.}
    \label{fig:MLCNarch}
\end{figure}

In 2019,  we introduced a  novel organization for the CapsNet named Multi-Lane CapsNet (MLCN)  with improved explainabily, performance and parallelization without decreasing accuracy or generalization power \cite{MLCN-SPL}. However, beyond just encoding the probability of a class, each vector also contains information to reconstruct the original image, with distinct dimensions of the vector representing different features of the image. With this in mind, we propose to split the original CapsNet architecture\footnote{source code in https://github.com/vandersonmr/lanes-capsnet} (Figure \ref{fig:MLCNarch}), dividing the PCs into independent sets called $lanes$. Each of these sets of PCs, a \lane, is responsible for one of the dimensions in the final digit capsules.

The number of PCs per \lane\ may vary, as well as the way they are computed. In the original CapsNet, two 2D convolutions are applied to the input image and then reshaped to produce the PCs. More convolutions may be applied, what we call the $depth$ of a \lane, or more filters can be used per convolution generating more capsules, what we call the $width$ of a \lane. Further, distinct dimensions of a final digit capsule can be generated by \lanes\ with different configurations (and thus distinct computational requirements).

There are two key advantages of this organization over the original CapsNet architecture. First, it allows parallelism of the execution, as each set of  PCs is constructed independently, improving performance and allowing training and deployment on distributed environments. Second, it improves the explainability of the network by associating different features of the image to each \lane.

%TODO: discuss how model paralelism vs data parallelism works

\subsection{CapsNet Parallelization}

A DNN can be paralyzed in different ways and normally finding the best way for a given network is a complex and hard task. The three most common are data parallelism, model parallelism and pipelining. 

The first, data parallelism, splits the data which is going to be computed. Basically, it divides the input batch into smaller batches for each computer unit and synchronizes it at the end of the batch. Although being very simple and straightforward, it can only scale increasing the batch size as dividing too much a small batch can result in small computation and frequent synchronization. And varying the batch size impacts in the accuracy, what can mean a trade-off between accuracy and speedup.

Another possibility is by splitting the network operations itself. However, it is not always trivial to find a good place to split the operations. Normally, if two operations which are data-dependent are split into two computation units, it will involve lots of communication. Moreover, implementing this kind of network division and communication in current frameworks is not trivial.

Lastly, but not less important, pipelining split the network into levels which can compute different data at the same time in a pipeline approach. It is normally the approach used in high-performance scenarios.

These are not the only techniques to distribute the training and inference of DNNs and they are not mutually exclusive and can be used together \cite{jia2018beyond}. Related to MLCN, we tested its capability of allowing easy model parallelism and compare it to the common approach that is data parallelism. Of course, for huge MLCN networks pipeline could also be used, but we focus on showing how being able to facilitate the use of model parallelism can bring many advantages mainly over when only using data parallelism. This same advantage can be easily extended by adding pipelining per lane, but it will remains for future work.

\section{Heterogeneous Distribution Problem}\label{sec:heteproblem}

\begin{figure*}[t!]
    \centering
    \includegraphics[width=0.85\textwidth]{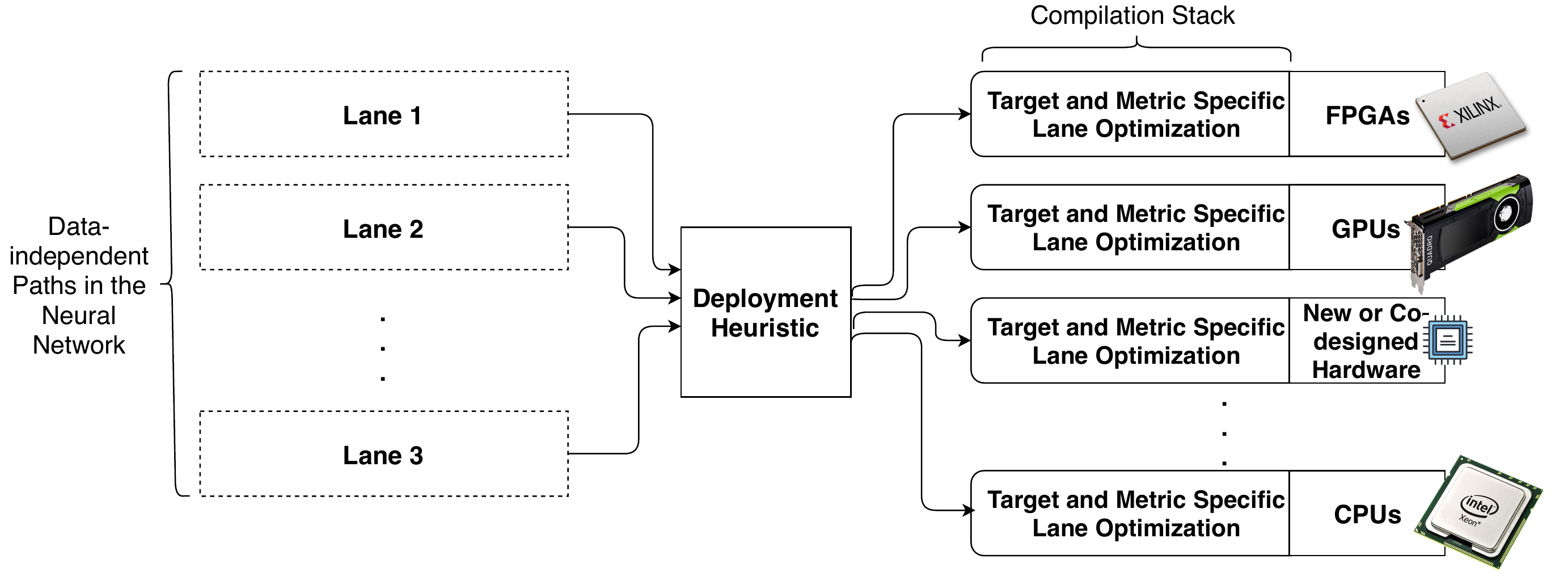}
    \caption{Multiple Neural Network \lanes\ can be trained in parallel using multiple HW even in heterogeneous scenarios..}
    \label{fig:autodeploy}
\end{figure*}

One of the main advantages of having data-independent \lanes\ is that these \lanes\ can be deployed separately in multiple accelerators. If we have multiple equal \lanes\ and multiple equal accelerators, deployment is as basic as dividing the \lanes\ equally over the HW resources, only being concerned with the communication cost involved. If in other cases we have \lanes\ with different shapes, characteristics and computational intensity or/and we have multiple accelerators with different characteristics or computational power, deployment becomes more involved. First, because it now involves load balancing the computational intensity of the \lanes\ and the computational power of the \lane\ and, second, because now there is also the chance to apply different optimizations for different pairs of HW and \lane. This scenario can be seen in Figure \ref{fig:autodeploy}, which shows how multiple \lanes\ can be deployed for different accelerators with different compilation stacks. 

Deciding where to execute each \lane\ or what optimizations to apply to each lane/hardware pair is not trivial. In this work, we present an approach to address the first problem using a deployment heuristic. We will address the second problem in future work.

\subsection{Heuristic Execution Cost for MLCN Lanes}

Finding, statically, the optimal solution  to deploy a \lane\ given a set of HW resources is a complex task. For example, aspects such as the version of the compiler being used or what other \lanes\ (and their characteristics) are being executed concurrently on the same HW can have a significant impact on the final performance. These are only two of the many aspects that can affect  performance. However, we observed in our experiments that, at least for MLCN, we do not need to have the exact final performance to make a good deployment decision. Our experiments have shown that  simple predictors can provide fair results.

We experiment running and taking the average execution time of 10 executions of MLCN \lanes\ with different width, size, and types, using three different NVIDIA GPUs (K80, P100, and V100). For the same number of parameters, independently of the GPU used, the performance displays a well-behaved pattern. It varies linearly when increasing the depth, quadratically when varying the width, and it was multiplied by a factor when changing the GPU.

Thus, for MLCN \lanes\ with NVIDIA GPUs and compilers, predicting the performance on a given HW substrate can be approximated by Equation \ref{eq:lanecost}, which achieves a 0.901 Pearson correlation with our experimental data. 

\begin{equation}
    lane_{cost} = (lane_{width})^2 \times lane_{depth} \times GPU_{speed}
    \label{eq:lanecost}
\end{equation}
\vspace{0.05cm}

The $GPU_{Speed}$ in equation \ref{eq:lanecost} is the speed factor of the GPU being used. It only has any significance when deploying to a heterogeneous set of GPUs and the $GPU_{Speed}$ constant for each GPU can be inferred by simply measuring the execution time of a tiny \lane\ in each GPU and normalizing it. This can be done before the execution and it has an insignificant cost in the final execution time. In the case of our experiments, we collect the $GPU_{Speed}$ by executing a 512x512 fully connected network with a small set of data. Normalized by K80, we used the following $GPU_{Speed}$s for M40, P100 and V100: 3.1, 4.2 and 6. 

\subsection{Load Balancing Algorithm}

We showed that we can  make good execution cost predictions for NVIDIA GPUs and MLCN \lanes. However, there is still the problem of how to deploy a set of \lanes\ with different sizes and widths  to a set of GPUs with different speeds. We can model this problem as a numerical set partition with $N$ bins,  each bin corresponding to a target GPU. The cost of each \lane\ being deployed (inserted into the bin) is equal to the \lane\ cost (Equation \ref{eq:lanecost}) multiplied by previous execution speed prediction on host HW via execution of a tiny \lane\ ($GPU_{Speed}$).

The numerical set partition problem is NP-Hard, but very good results can be achieved using heuristic/approximative algorithms and it can even be solved in pseudo-polynomial time using dynamic programming making it one of ``The Easiest Hard Problem" \cite{hayes2002computing, korf2009multi}. One of such heuristics that achieved good results and is very simple to implement is the greedy partition which always inserts the remaining \lane\ with the largest cost in the emptiest bin. Algorithm \ref{al:greedy} shows this greedy algorithm including the pre-execution used to calculate the $GPU_{Speed}$. 

\begin{algorithm}[ht!]
\small
\begin{lstlisting}[language=Python]
if using heterogeneous HW:
  for each HW:
    execute a tiny lane
    GPUSpeed[i] = runtime of the tiny lane
    GPUSpeed[i] = GPUSpeed[i]/smallest(GPUSpeed)

def GreedyPartition(lanes, NumGPUs,GPUSpeed):
  GPUTasks = [[] for i in range(NumGPUs)]
  for lane in reverse sorted lanes:
    sort GPUTasks by GPUTasks[i][j]*GPUSpeed[i] 
    GPUTasks[0].append(lane)
  return GPUTasks
\end{lstlisting}
\caption{Greedy Parition Algorithm}
\label{al:greedy}
\end{algorithm}

\section{Experimental Setup}\label{sec:expsetup}

In our experiments, we used machines from Google Cloud. All virtual machines instantiated had 24 vCPUs with 50GB of RAM and a default network interface. We used different GPU setups, including NVIDIA Tesla M40, K80, P100 and V100 all with CUDA 10.0, Intel MKL-DNN and Tensorflow 1.13.1.

The results and experiments that we explore did not show sensitivity to the input data set (tested with MNIST, CIFAR10, and others) and we chose to use the MNIST data set. Execution time was measured by executing 10 MNIST epochs, excluding the first, and using the average time for the others. All results had a very small variation. The execution time between epochs had always a very similar value. Thus, for simplicity, we present averages.

Thus, in this work we tested four configurations for the CapsNet parallelization, as follow:

\begin{itemize}
    \item Original with Data Parallelism (\textbf{baseline or base}): we simply used the original concept of CapsNet for the MNIST dataset parallelized using Keras data parallelism support.
    \item MLCN with Data Parallelism (\textbf{mlcn-data}): we used the same approach as in the baseline (Keras data parallelism), but with the MLCN organization.
    \item MLCN with Model Parallelism (\textbf{mlcn-model}): we parallelize the execution by executing each \lane\ on different GPUs. When using multiple machines, we used Horovod MPI framework to do handle the communication.
\end{itemize}

\section{Experimental Results}\label{sec:expresult}

\subsection{MLCN Scalability}

To understand how each approach to the parallelization of CapsNet scales, we studied their performance with 1, 2, 4 and 8 NVIDIA Tesla K80 GPUs. 

\begin{figure}[H]
\centering
\includegraphics[width=0.85\linewidth]{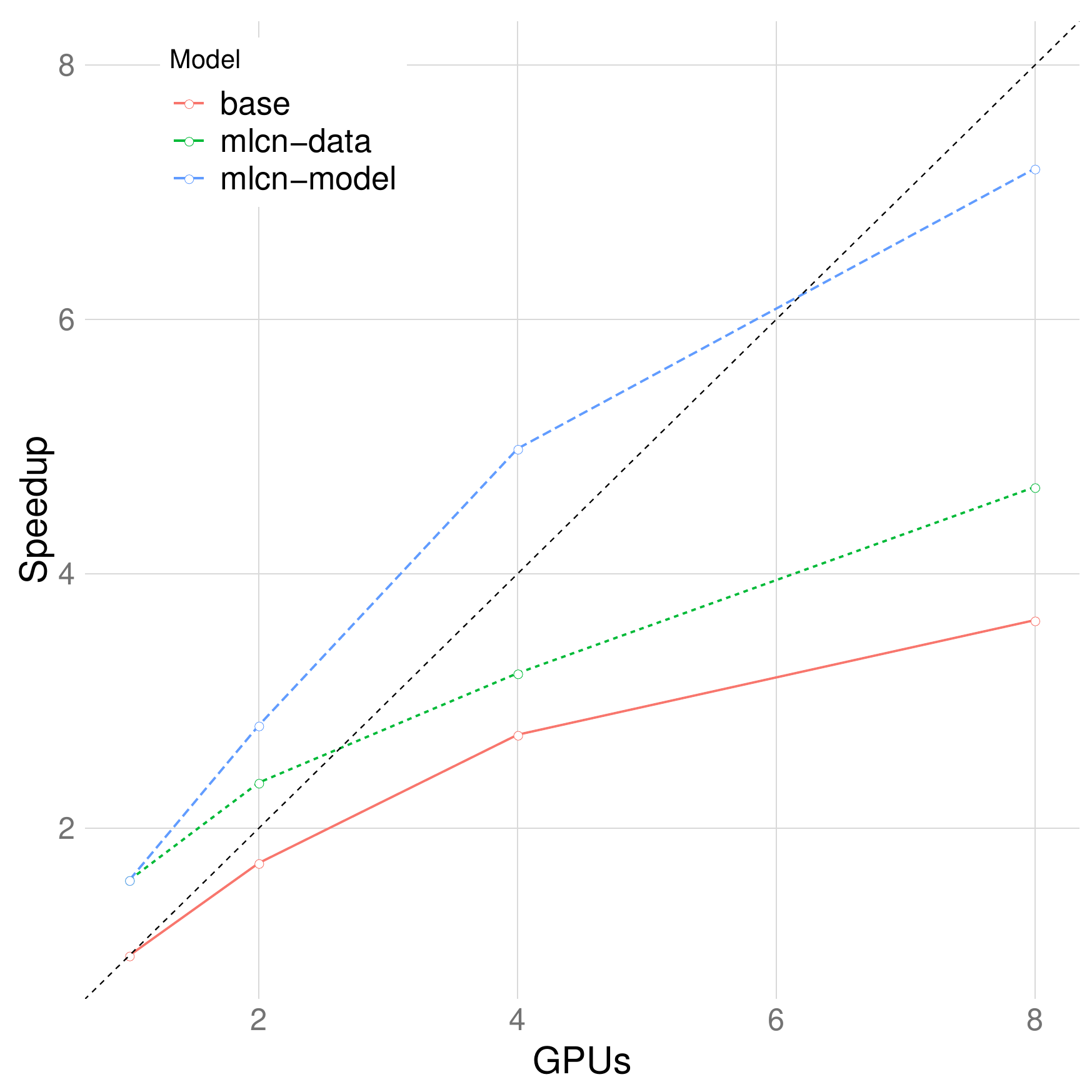}
\caption{speedup of the three parallelization approaches: baseline with data parallelism (base), MLCN with data parallelism (mlcn-data) and MLCN with model parallelism (mlcn-model). All speedup are relative to the baseline with one GPU.}
\label{fig:speedupall}
\end{figure}

The graph in Figure \ref{fig:speedupall} shows the performance comparison between the base (baseline), mlcn-data and mlcn-model. MLCN is faster than the baseline even in a single GPU, as reported earlier. However, it is interesting to notice that the advantage does not increase when scaling to more GPUs with data parallelization, as the speedup difference between mlcn-data and baseline remained constant. This suggests that the reorganization proposed by  MLCN does not improve  scaling via data parallelism. However, the same is not true for  model-parallelism. In this case Mlcn-model has a visible advantage, scaling with higher efficiency and achieving a near  7.18 speedup with 8 GPUs over the single GPU baseline. Thus, MLCN  not only is faster  than the original CapsNet (baseline) but, because it allows model-parallelism, it scales more efficiently.

\subsection{Impact of Batch Size}

\begin{figure}[H]
\centering
    \subfloat[\label{fig:datascale}Baseline using data-parallelism for different mini batch sizes]{\includegraphics[width=0.85\linewidth]{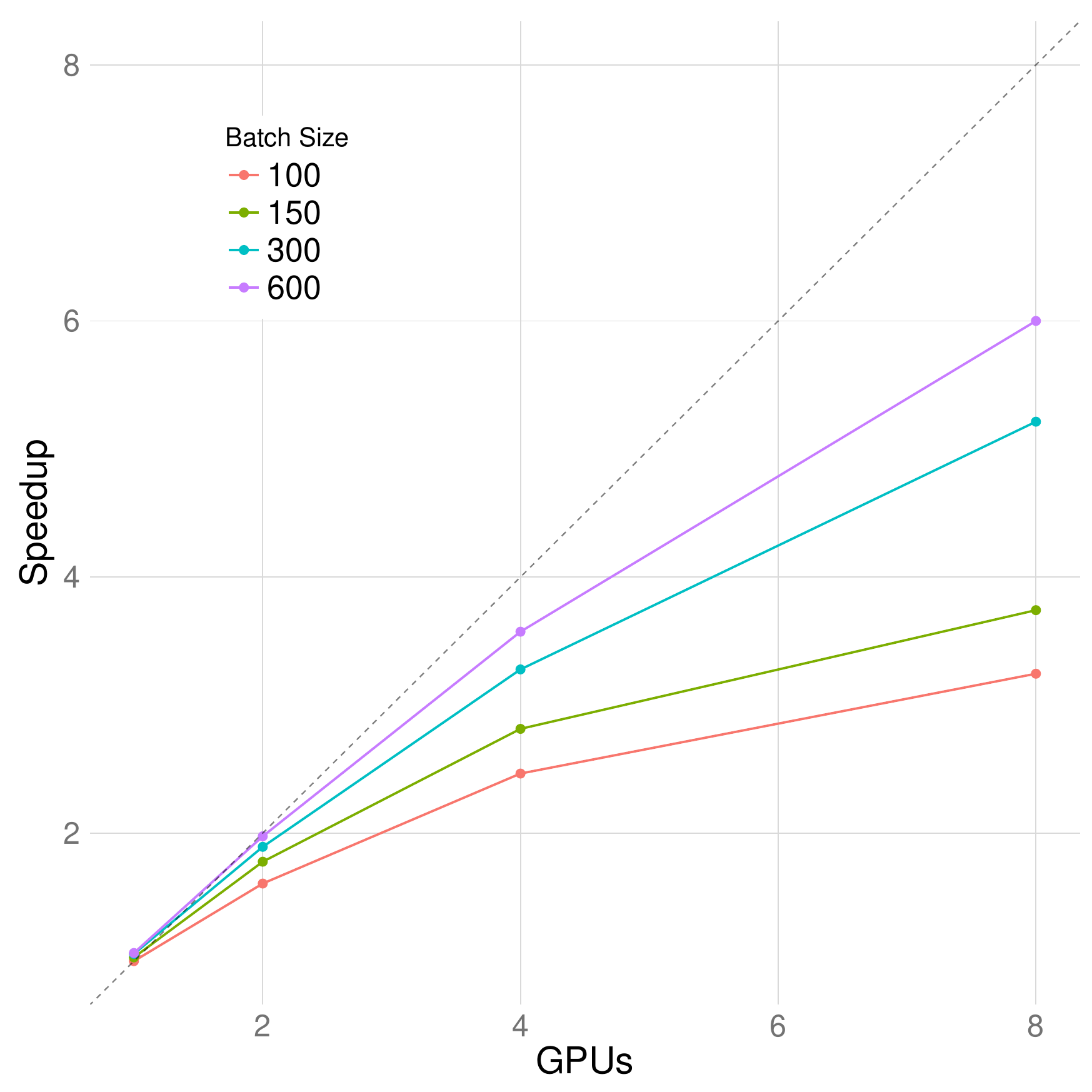}} \vspace{1cm}
    \\
    \subfloat[\label{fig:mlcnscale}MLCN using model-parallelism for different mini batch sizes]{\includegraphics[width=0.85\linewidth]{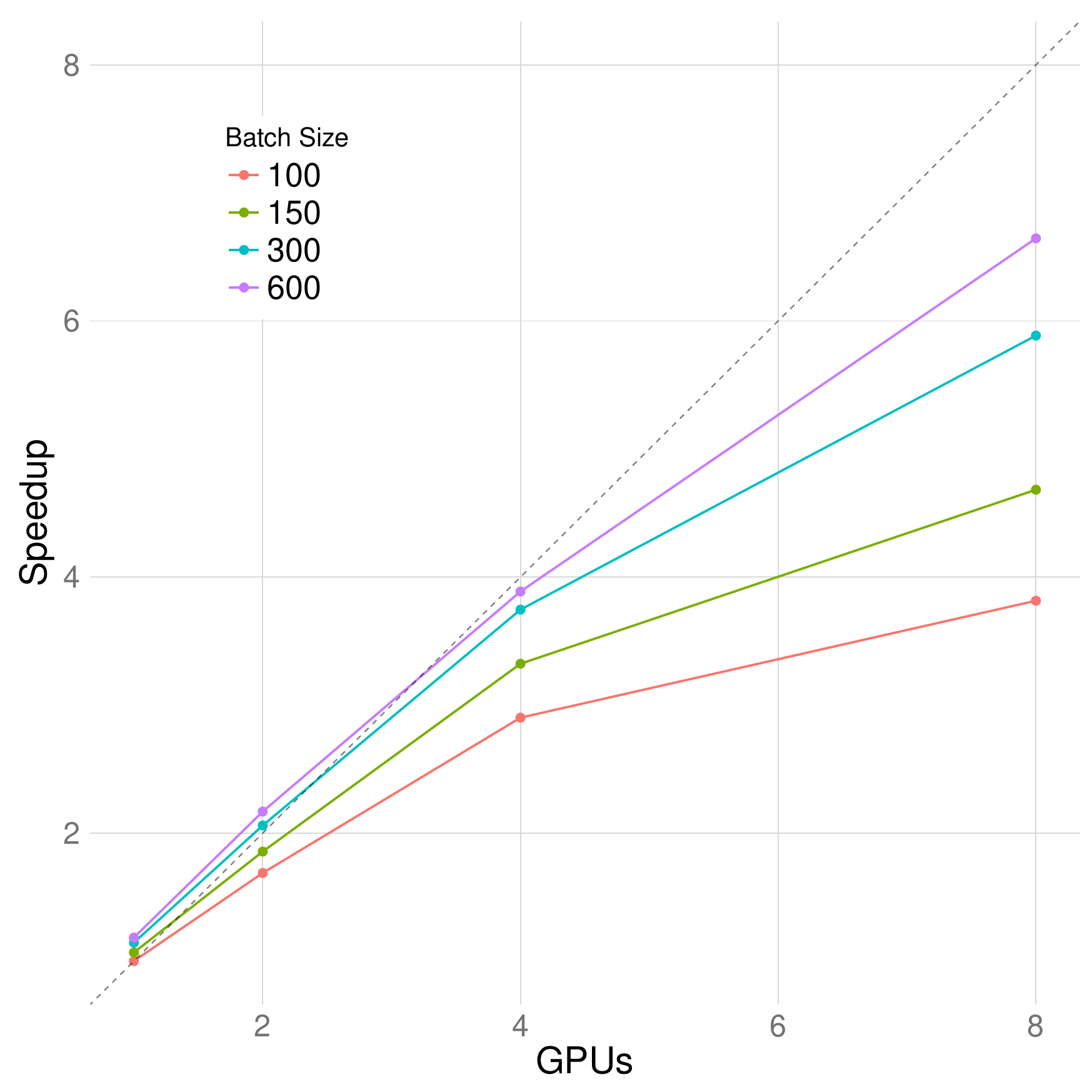}} 
    \label{fig:scaleMLCNxbase}
    \caption{MLCN and baseline scalability for 1, 2, 4 and 8 NVIDIA K80 GPUs using Google Cloud VM with 24 vCPUs and 90GB of RAM.}
\end{figure}

The size of the minibatch, or batch size, has a significant impact on the performance of a DNN as more computation/communication is available, enabling a more efficient use of the HW. The batch size has a significant impact on data parallelism performance as more data/computation is available to be divided among the GPUs. To study   the advantage of MLCN over the data parallelism method we tested both approaches with 100, 150, 300 and 600 batch sizes. The graphs in Figures \ref{fig:datascale} and \ref{fig:mlcnscale} show the speedup  versus a single GPU with a 100-sized batch size. In both cases we  observe similar efficiency  as batch size grows. So, for different batch sizes the relative advantage of  MLCN with model parallelism stays the same, as increasing the batch size equally increases  the efficiency of  data and model parallelism approaches.

\begin{figure}[H]
\centering
\includegraphics[width=0.85\linewidth]{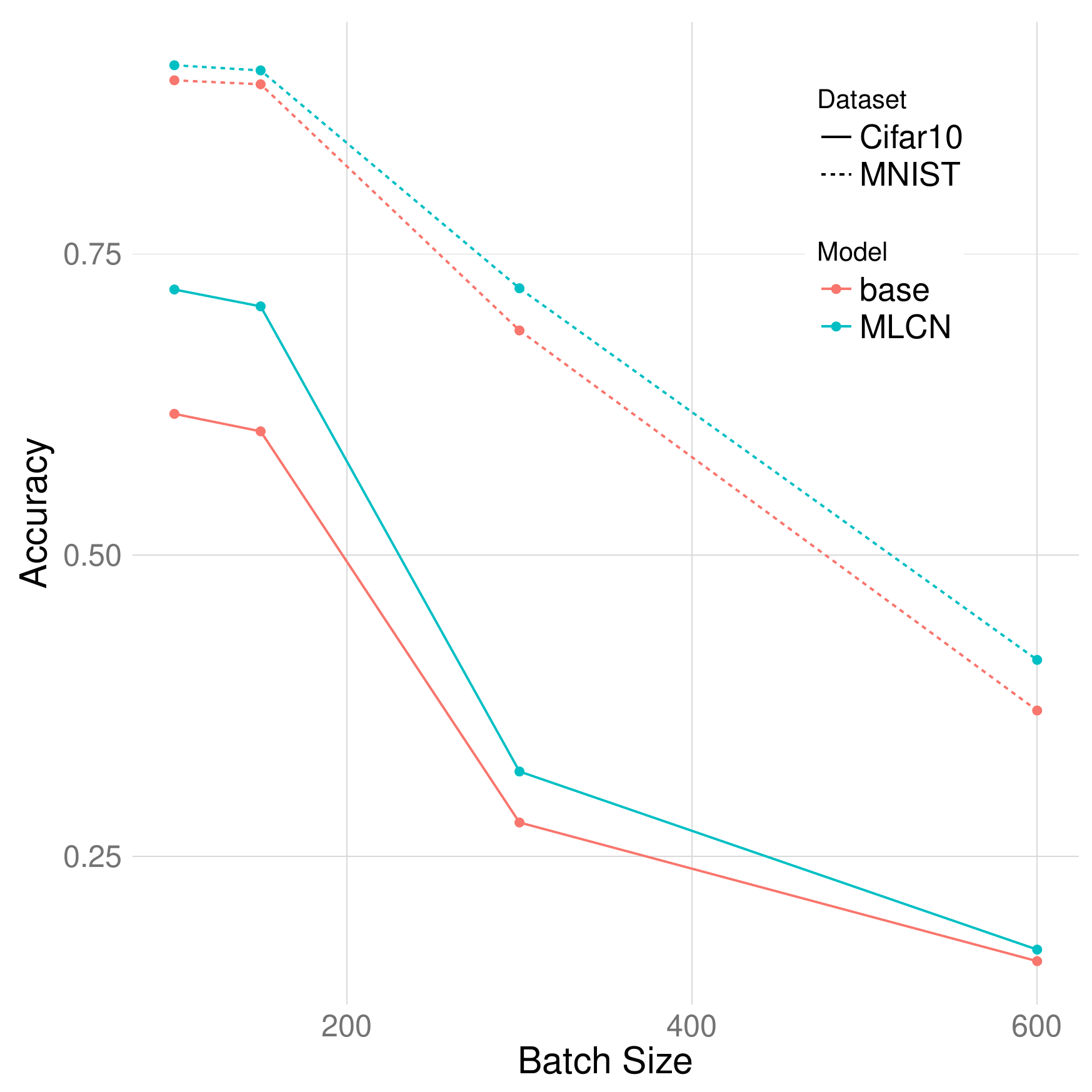}
\caption{validation accuracy impact when increasing the training batch size for the baseline and MLCN in the Cifar10 and MNIST datasets.}
\label{fig:batch-size}
\end{figure}

We also studied the impact of  batch size on both baseline and MLCN accuracy, shown in  Figure \ref{fig:batch-size}. Increasing  batch sizes have a significant impact on the accuracy in both cases.  
The magnitude of this impact is related to the dataset as shown by the differences between MNIST and Cifar10 results. Thus, as model parallelism has  better performance and scalability with smaller batch sizes (Figures \ref{fig:datascale} and \ref{fig:mlcnscale}), model parallelism has the advantage of scaling without the need to trade accuracy for efficiency.

\subsection{Impact of Lanes Characteristics}

\begin{figure}[H]
\centering
    \subfloat[\label{fig:scalesize}MLCN using model-parallelism with mini batch width of 150 and varying the width of the \lanes.]{\includegraphics[width=0.83\columnwidth]{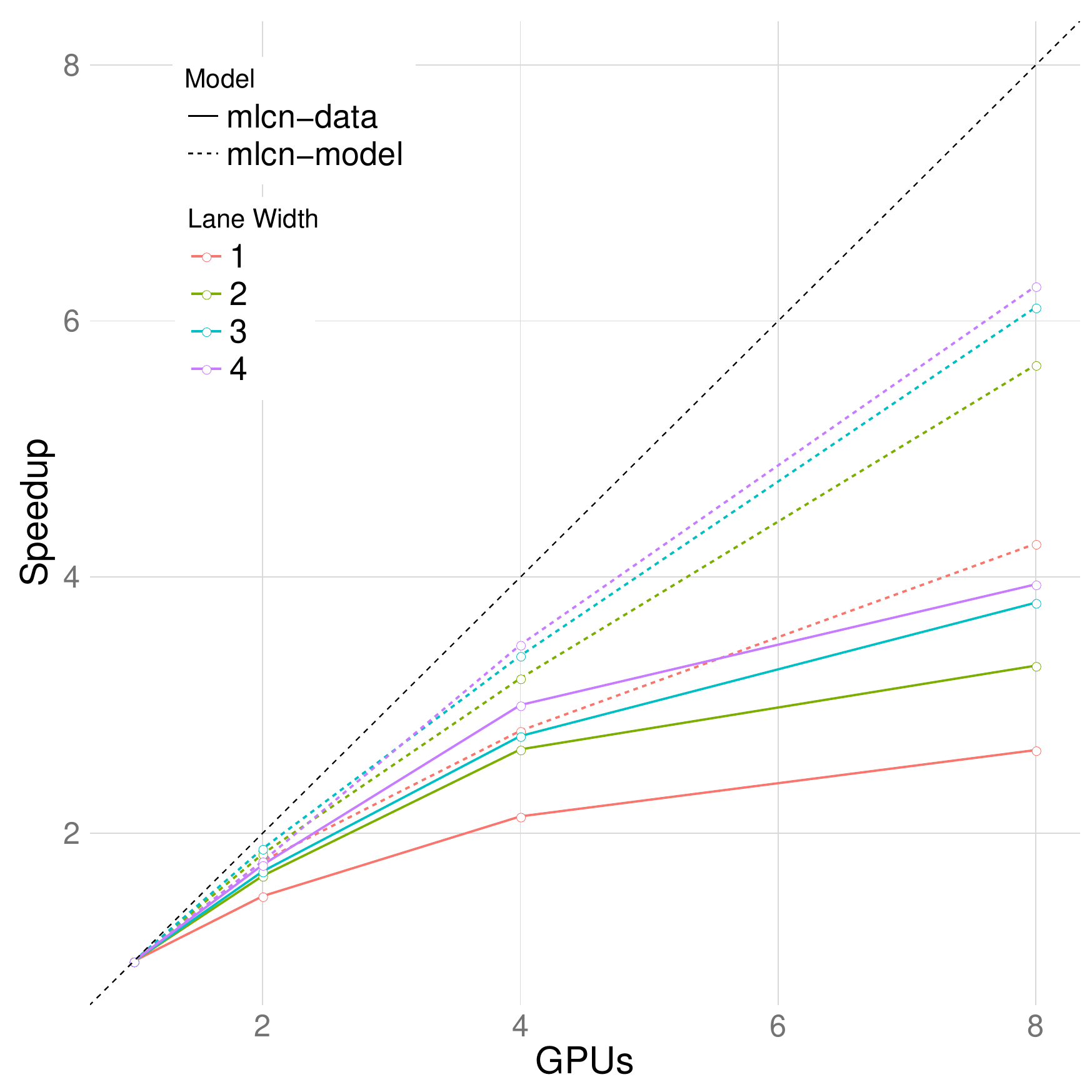}}
    \hspace{0.3cm}
    \subfloat[\label{fig:scaledepth}MLCN using model-parallelism with mini batch width of 150 and varying the size of the \lanes.]{\includegraphics[width=0.83\columnwidth]{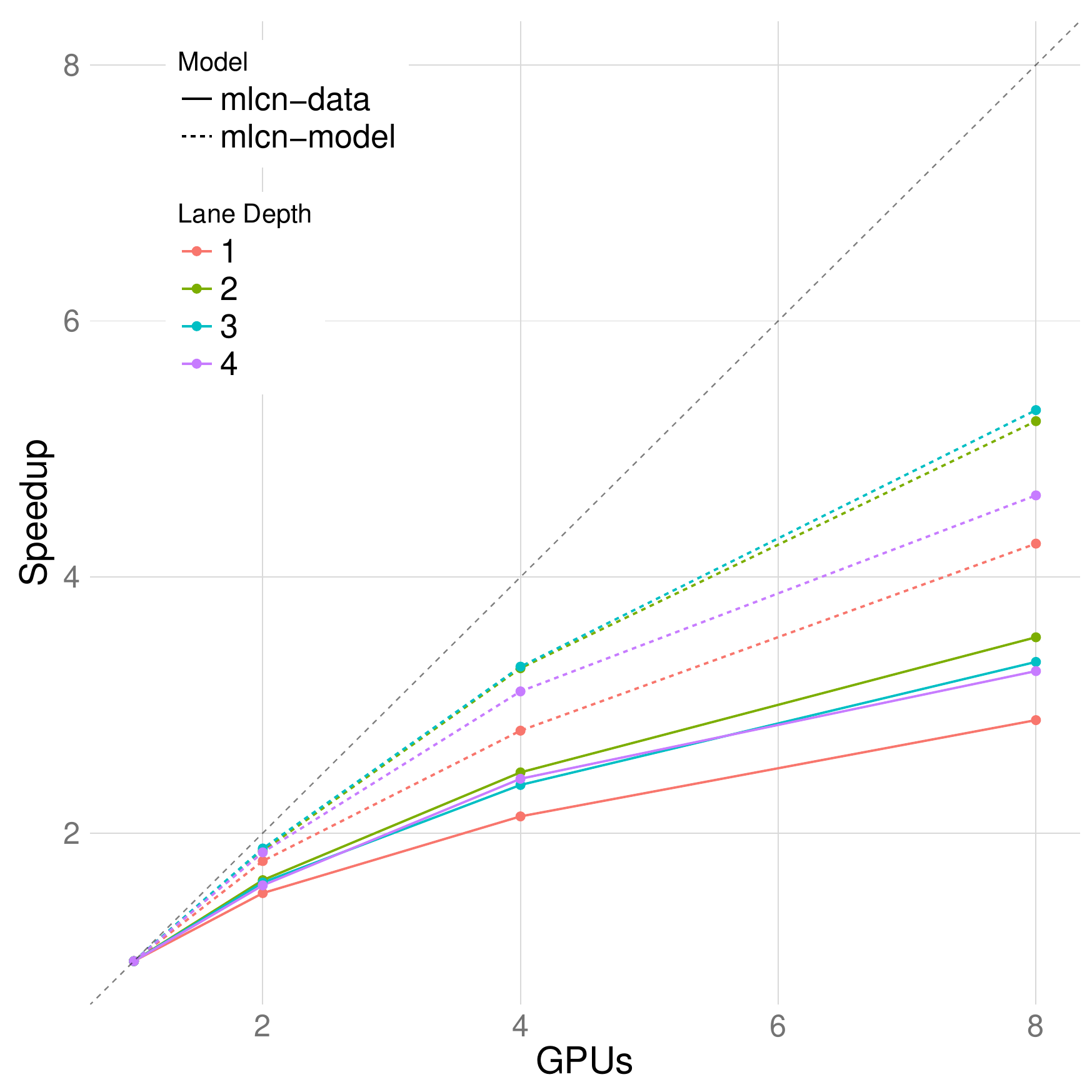}}
    \hspace{0.3cm}
    \subfloat[\label{fig:scalenumber}MLCN using model-parallelism with batch size of 150 and varying the number of \lanes.]{\includegraphics[width=0.83\columnwidth]{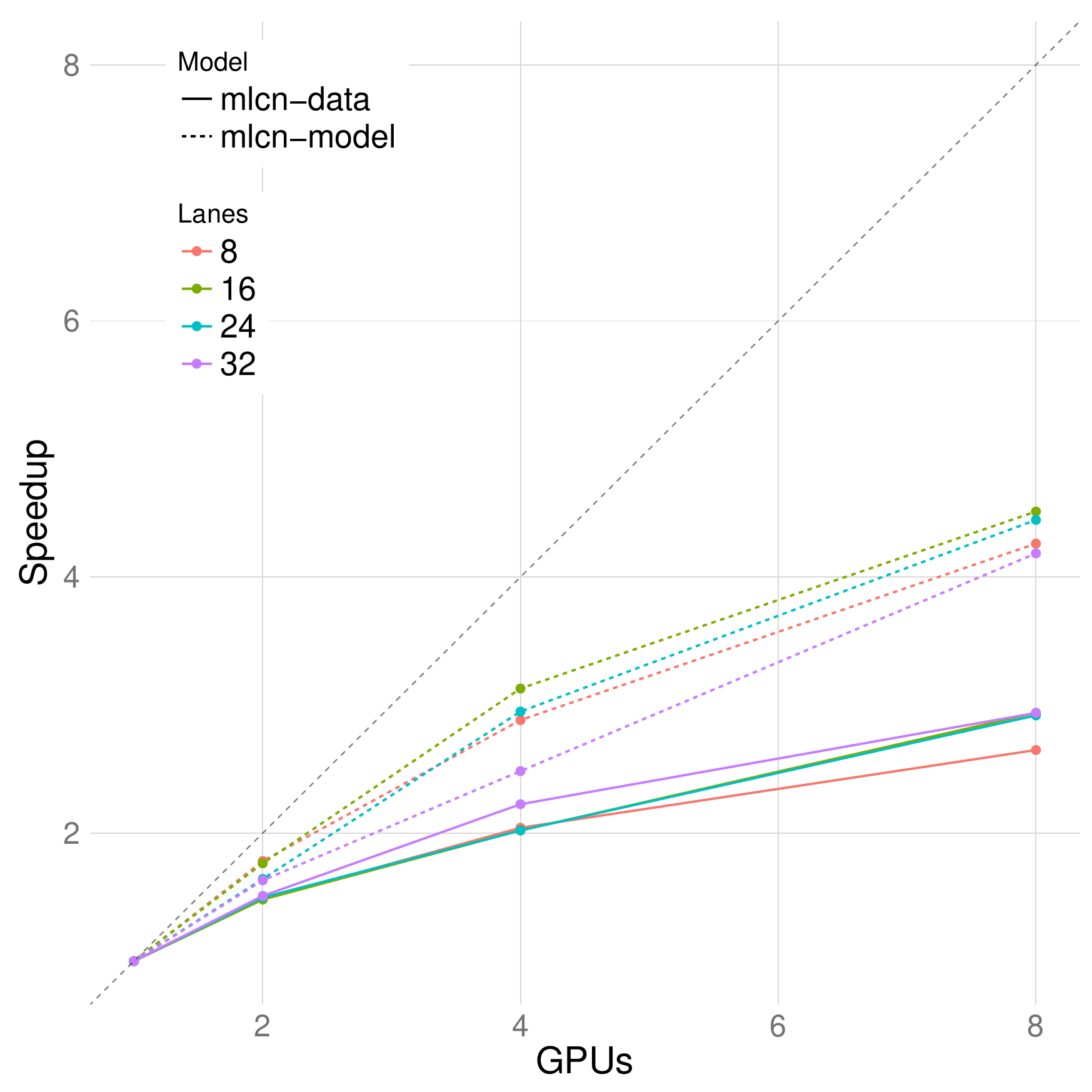}} 
    \caption{scalability variance with different \lanes\ configurations.}
    \label{fig:scale}
\end{figure}

The previous results explored the suitability  of  MLCN and its model parallelization. We also explore also how the characteristics of the MLCN \lanes\ can affect  performance and scalability by  varying the three main hyperparameters in  MLCN \lanes: their width, depth and the quantity. The results are shown, respectively, in Figures \ref{fig:scalesize}, \ref{fig:scaledepth} and \ref{fig:scalenumber}.

The width and  depth of  \lanes\ has a direct impact on the number of parameters per \lane\ and, consequently, the amount of computation per \lane. With more computation per \lane, the  efficient  use of multiple GPUs becomes advantageous. This is  shown in Figures \ref{fig:scalesize} and \ref{fig:scaledepth} as larger lanes increase efficiency. However, increasing the width had a much more significant increase in efficiency, at similar increase in number of parameters. This indicates that, besides the number of parameters, the type of computation affects performance. In the case of MLCN \lanes\, wider \lanes\ result in  better performance than deeper \lanes\ with the same number of parameters.

Another interesting point was the fact that increasing the number of lanes did not significantly increase   performance, as shown in Figure \ref{fig:scalenumber}. Even though increasing the number of lanes also increases the amount of computation available between batches, there is an overhead of having these computations separable. So, having several \lanes\ in one GPU is less efficient than having a single extremely large \lane.

\subsection{Heterogeneous Lanes and GPUs}

\begin{figure}[H]
\centering
\includegraphics[width=0.85\linewidth]{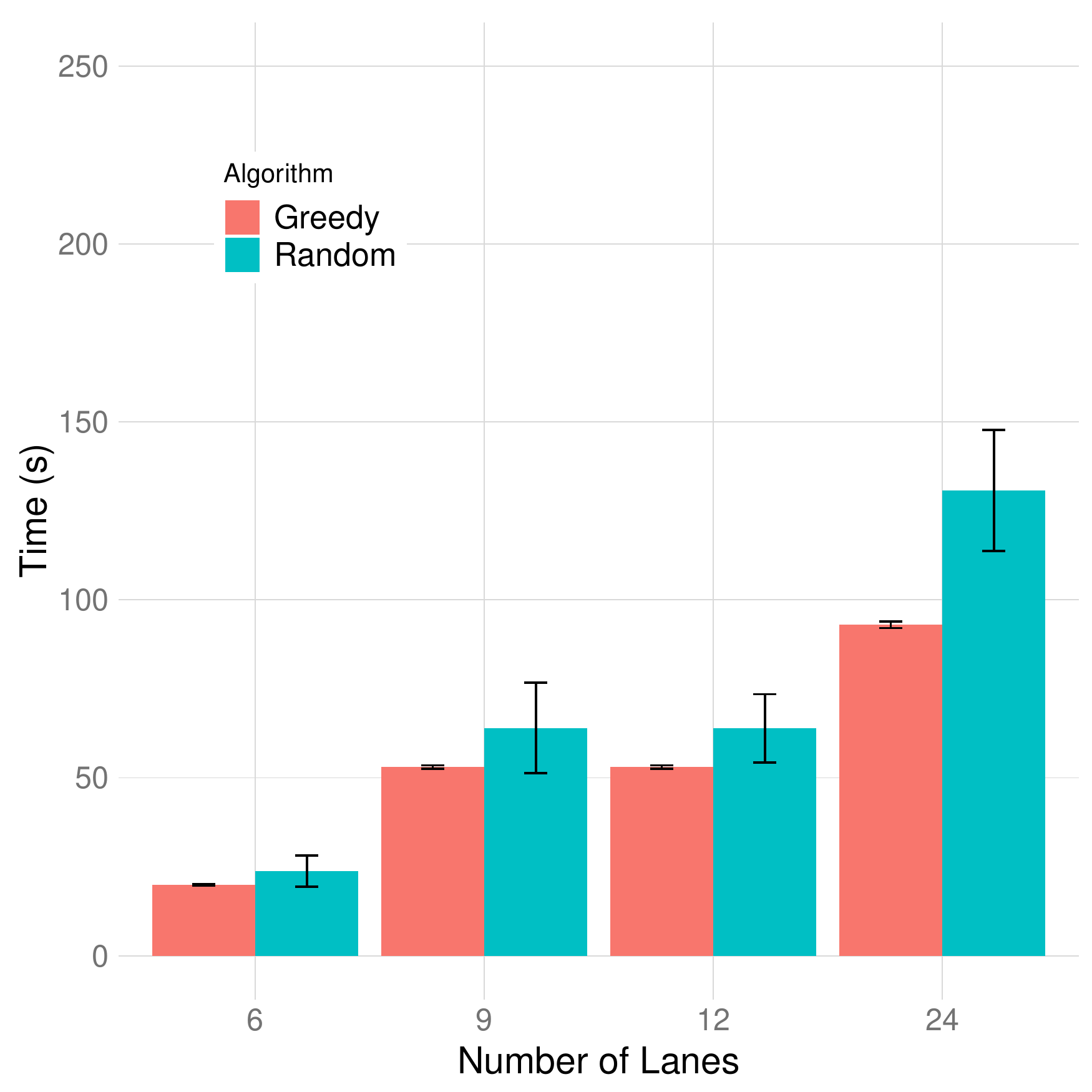}
\caption{average execution time (executed 10 times) of heterogeneous \lanes\ running on four K80 NVIDIA GPUs with a random and a greedy partition of \lanes\ execution distribution. All \lanes\ varying on width and depth.}
\label{fig:hetelanes}
\end{figure}

One interesting  observation about MLCN is that having \lanes\ with different characteristics, such as \lanes\ with different sizes and depths, increases the generality of the network. A similar result was reported by Canqun \textit{et al.} \cite{xiang2018ms} with the MS-CapsNet organization. However, as  discussed in Section \ref{sec:heteproblem}, deploying \lanes\ in multiple GPUs when the \lanes\ have different computational footprint can be challenging. To study a proposed heuristic to deploy \lanes\ with different widths and depths, we tested 4 MLCN networks with 6,  9, 12 and 24 \lanes. Each \lane\ may have  pairs of depth and width values ranging from 1 to 5. As shown in Figure \ref{fig:hetelanes}, we obtain a smaller execution time with our heuristic than when na\"ively randomly distributing the \lanes\ between the GPUs. The advantage increases with the number of \lanes, showing that,  the larger the number of \lanes\, the harder it is to randomly find a good distribution. Notice that the time accounted for the greedy heuristic includes the (almost insignificant) time to run the heuristic.

%\vspace{1cm}

\subsection{Heterogeneous Lanes with Heterogeneous GPU}

\begin{figure}[H]
\centering
\includegraphics[width=0.85\linewidth]{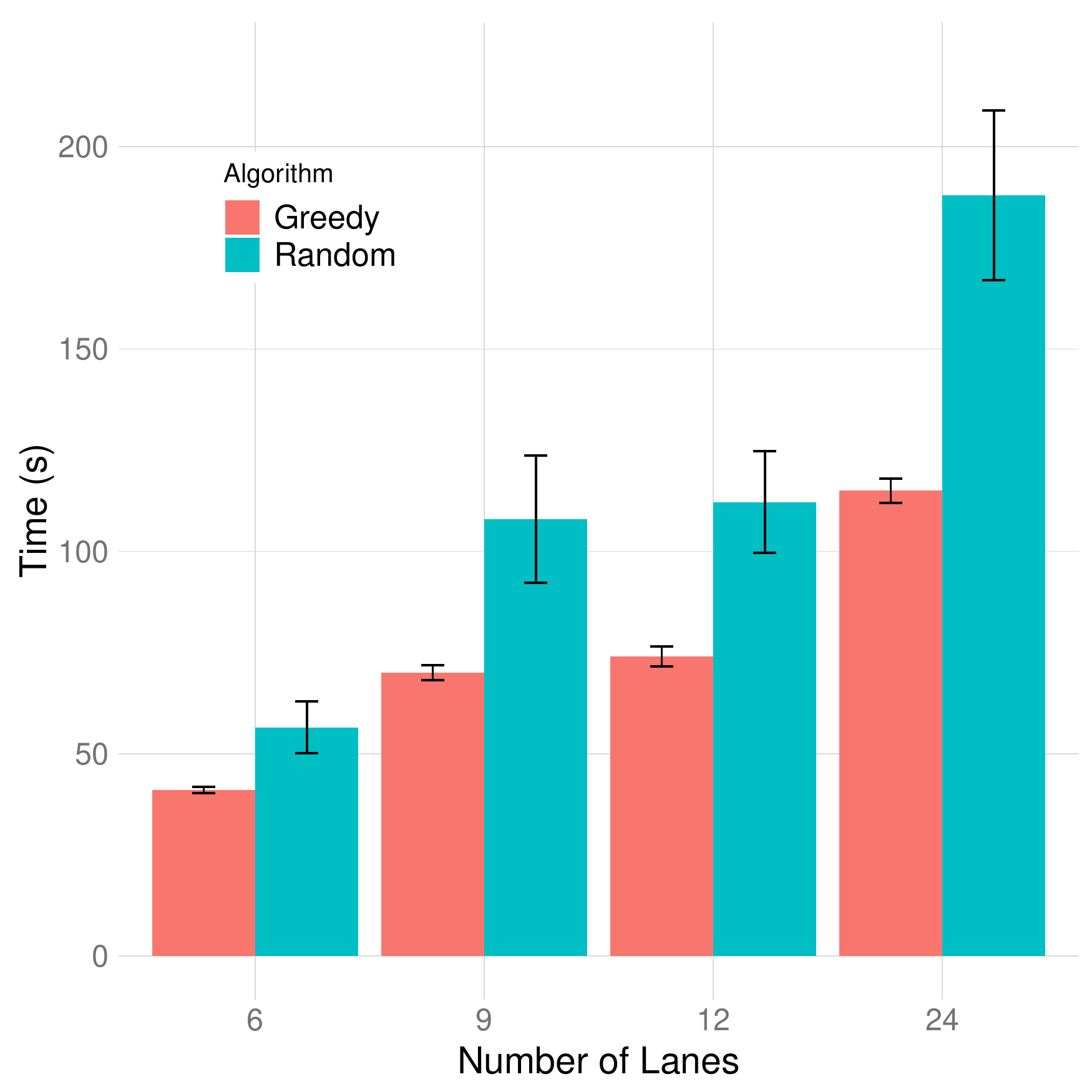}
\caption{average execution time (executed 10 times) of heterogeneous \lanes\ running on one K80, one P100, one V100, and one M40 NVIDIA GPU in multiple machines communicating using MPI with a random and a greedy partition of \lanes\ execution distribution. All \lanes\ varying on width and depth.}
\label{fig:hetegpus}
\end{figure}

More than having heterogeneous \lanes\ we also tested a scenario with heterogeneous accelerators. Rather than 4 NVIDIA Tesla K80, we deployed  four systems each  with a different GPU: one M40, one K80, one P100, and one V100. The results are in Figure \ref{fig:hetegpus}. For total execution time, there was a significant increase because of network communication between the systems. Moreover, the difference between  random deployment and our greedy heuristic becomes larger, showing that for more complex the scenarios with many \lanes\ or heterogeneous HW, it is  key to deploy the computation carefully.

%\vspace{1cm}

\section{Conclusion}\label{sec:conclusion}

The Multi-lane CapsNet (MLCN) is a novel organization for the CapsNet network which is shown to achieve better accuracy  with more efficient HW utilization. Further,  MLCN allows  model parallelization by running the \lanes\ in parallel. In this work, we analyze and measure the advantages of this new parallelization scheme of the CapsNet when compared to the usual data parallelism.

We find that MLCN is faster than the original CapsNet and it scales better with  model parallelism being almost 2x more efficient, even with small batch sizes. We also explored the impact of different \lane\ configurations on performance and scalability, showing that wider \lanes\ usually achieve higher HW  efficiency.

Finally, we found that when parallelizing MLCN with \lanes\ with different characteristics (or when deploying in machines with different accelerators), load balance is a key factor to reaching good performance. 
We proposed a greedy algorithm to deploy \lanes\ in these scenarios and we found that it can be up to 50\% more efficient than the na\"ive random deployment.

\bibliographystyle{IEEEtran}
\bibliography{ref}

\begin{thebibliography}{10}
\providecommand{\url}[1]{#1}
\csname url@samestyle\endcsname
\providecommand{\newblock}{\relax}
\providecommand{\bibinfo}[2]{#2}
\providecommand{\BIBentrySTDinterwordspacing}{\spaceskip=0pt\relax}
\providecommand{\BIBentryALTinterwordstretchfactor}{4}
\providecommand{\BIBentryALTinterwordspacing}{\spaceskip=\fontdimen2\font plus
\BIBentryALTinterwordstretchfactor\fontdimen3\font minus
  \fontdimen4\font\relax}
\providecommand{\BIBforeignlanguage}[2]{{%
\expandafter\ifx\csname l@#1\endcsname\relax
\typeout{** WARNING: IEEEtran.bst: No hyphenation pattern has been}%
\typeout{** loaded for the language `#1'. Using the pattern for}%
\typeout{** the default language instead.}%
\else
\language=\csname l@#1\endcsname
\fi
#2}}
\providecommand{\BIBdecl}{\relax}
\BIBdecl

\bibitem{huang2018gpipe}
Y.~Huang, Y.~Cheng, D.~Chen, H.~Lee, J.~Ngiam, Q.~V. Le, and Z.~Chen, ``Gpipe:
  Efficient training of giant neural networks using pipeline parallelism,''
  \emph{arXiv preprint arXiv:1811.06965}, 2018.

\bibitem{mehta2018high}
R.~Mehta, Y.~Huang, M.~Cheng, S.~Bagga, N.~Mathur, J.~Li, J.~Draper, and
  S.~Nazarian, ``High performance training of deep neural networks using
  pipelined hardware acceleration and distributed memory,'' in \emph{2018 19th
  International Symposium on Quality Electronic Design (ISQED)}.\hskip 1em plus
  0.5em minus 0.4em\relax IEEE, 2018, pp. 383--388.

\bibitem{ben2018demystifying}
T.~Ben-Nun and T.~Hoefler, ``Demystifying parallel and distributed deep
  learning: An in-depth concurrency analysis,'' \emph{arXiv preprint
  arXiv:1802.09941}, 2018.

\bibitem{jia2018beyond}
Z.~Jia, M.~Zaharia, and A.~Aiken, ``Beyond data and model parallelism for deep
  neural networks,'' \emph{arXiv preprint arXiv:1807.05358}, 2018.

\bibitem{MLCN-SPL}
V.~M. do~Rosario ; Edson Borin ; Mauricio~Breternitz, ``The multi-lane capsule
  network,'' \emph{IEEE Signal processing letters}, vol.~26, pp. 1006--1010,
  2019.

\bibitem{chollet2017xception}
F.~Chollet, ``Xception: Deep learning with depthwise separable convolutions,''
  in \emph{Proceedings of the IEEE conference on computer vision and pattern
  recognition}, 2017, pp. 1251--1258.

\bibitem{szegedy2017inception}
C.~Szegedy, S.~Ioffe, V.~Vanhoucke, and A.~A. Alemi, ``Inception-v4,
  inception-resnet and the impact of residual connections on learning,'' in
  \emph{Thirty-First AAAI Conference on Artificial Intelligence}, 2017.

\bibitem{hinton2011transforming}
G.~E. Hinton, A.~Krizhevsky, and S.~D. Wang, ``Transforming auto-encoders,'' in
  \emph{International Conference on Artificial Neural Networks}.\hskip 1em plus
  0.5em minus 0.4em\relax Springer, 2011, pp. 44--51.

\bibitem{sabour2017dynamic}
S.~Sabour, N.~Frosst, and G.~E. Hinton, ``Dynamic routing between capsules,''
  in \emph{Advances in neural information processing systems}, 2017, pp.
  3856--3866.

\bibitem{shahroudnejad2018improved}
A.~Shahroudnejad, A.~Mohammadi, and K.~N. Plataniotis, ``Improved
  explainability of capsule networks: Relevance path by agreement,''
  \emph{arXiv preprint arXiv:1802.10204}, 2018.

\bibitem{jaiswal2018capsulegan}
A.~Jaiswal, W.~AbdAlmageed, Y.~Wu, and P.~Natarajan, ``Capsulegan: Generative
  adversarial capsule network,'' in \emph{European Conference on Computer
  Vision}.\hskip 1em plus 0.5em minus 0.4em\relax Springer, 2018, pp. 526--535.

\bibitem{ren2018compositional}
H.~Ren and H.~Lu, ``Compositional coding capsule network with k-means routing
  for text classification,'' \emph{arXiv preprint arXiv:1810.09177}, 2018.

\bibitem{jimenez2018capsule}
A.~Jim{\'e}nez-S{\'a}nchez, S.~Albarqouni, and D.~Mateus, ``Capsule networks
  against medical imaging data challenges,'' in \emph{Intravascular Imaging and
  Computer Assisted Stenting and Large-Scale Annotation of Biomedical Data and
  Expert Label Synthesis}.\hskip 1em plus 0.5em minus 0.4em\relax Springer,
  2018, pp. 150--160.

\bibitem{mobiny2018fast}
A.~Mobiny and H.~Van~Nguyen, ``Fast capsnet for lung cancer screening,''
  \emph{arXiv preprint arXiv:1806.07416}, 2018.

\bibitem{kim2018capsule}
Y.~Kim, P.~Wang, Y.~Zhu, and L.~Mihaylova, ``A capsule network for traffic
  speed prediction in complex road networks,'' in \emph{2018 Sensor Data
  Fusion: Trends, Solutions, Applications (SDF)}.\hskip 1em plus 0.5em minus
  0.4em\relax IEEE, 2018, pp. 1--6.

\bibitem{mukhometzianov2018capsnet}
R.~Mukhometzianov and J.~Carrillo, ``Capsnet comparative performance evaluation
  for image classification,'' \emph{arXiv preprint arXiv:1805.11195}, 2018.

\bibitem{xiang2018ms}
C.~Xiang, L.~Zhang, Y.~Tang, W.~Zou, and C.~Xu, ``Ms-capsnet: A novel
  multi-scale capsule network,'' \emph{IEEE Signal Processing Letters},
  vol.~25, no.~12, pp. 1850--1854, 2018.

\bibitem{path2019}
M.~Amer and T.~Maul, ``Path capsule networks,'' in \emph{preprint, arxiv.},
  2019.

\bibitem{hayes2002computing}
B.~Hayes, ``Computing science: The easiest hard problem,'' \emph{American
  Scientist}, vol.~90, no.~2, pp. 113--117, 2002.

\bibitem{korf2009multi}
R.~E. Korf, ``Multi-way number partitioning,'' in \emph{Twenty-First
  International Joint Conference on Artificial Intelligence}, 2009.

\end{thebibliography}
\end{document}